\def\year{2021}\relax
\documentclass[letterpaper]{article} 
\usepackage{aaai21}  
\usepackage{times}  
\usepackage{helvet} 
\usepackage{courier}  
\usepackage[hyphens]{url}  
\usepackage{graphicx} 
\usepackage{graphicx}  
\usepackage{natbib}  
\usepackage{caption} 

\usepackage{amsmath}  
\usepackage{multirow}
\usepackage{subfig}
\newcommand{\tabincell}[2]{\begin{tabular}{@{}#1@{}}#2\end{tabular}}

\title{Delayed Feedback Modeling for the Entire Space Conversion Rate Prediction}
	\author {
	Yanshi Wang,\textsuperscript{\rm 1}
	Jie Zhang, \textsuperscript{\rm 1}
	Qing Da, \textsuperscript{\rm 1}
	Anxiang Zeng \textsuperscript{\rm 1} \\
}
\affiliations { \textsuperscript{\rm 1} Alibaba Group \\
     \{yanshi.wys,suowei.zj,daqing.dq,renzhong\}@alibaba-inc.com
}

\begin{document}
	\maketitle
	\begin{abstract}
		Estimating post-click conversion rate (CVR) accurately is crucial in E-commerce. However, CVR prediction usually suffers from three major challenges in practice: i) data sparsity: compared with impressions, conversion samples are often extremely scarce; ii) sample selection bias: conventional CVR models are trained with clicked impressions while making inference on the entire space of all impressions; iii) delayed feedback: many conversions can only be observed after a relatively long and random delay since clicks happened, resulting in many false negative labels during training. Previous studies mainly focus on one or two issues while ignoring the others. In this paper, we propose a novel neural network framework ESDF to tackle the above three challenges simultaneously. Unlike existing methods, ESDF models the CVR prediction from a perspective of entire space, and combines the advantage of user sequential behavior pattern and the time delay factor. Specifically, ESDF utilizes sequential behavior of user actions on the entire space with all impressions to alleviate the sample selection bias problem. By sharing the embedding parameters between CTR and CVR networks, data sparsity problem is greatly relieved. Different from conventional delayed feedback methods, ESDF does not make any special assumption about the delay distribution. We discretize the delay time by day slot and model the probability based on survival analysis with deep neural network, which is more practical and suitable for industrial situations. Extensive experiments are conducted to evaluate the effectiveness of our method. To the best of our knowledge, ESDF is the first attempt to unitedly solve the above three challenges in CVR prediction area. More than that, we release a sampling version of industrial dataset to enable the future research, which is the first public dataset that comprises impression, click and delayed conversion labels for CVR modeling.
	\end{abstract}
	
	\section{Introduction}
	\noindent  Estimating click-through rate (CTR) and conversion rate (CVR) accurately plays a vital role in E-commerce search and recommendation systems. It helps discover valuable products and better understand users' purchasing intention. Due to the huge commercial values, much efforts have been devoted to designing intelligent CTR and CVR algorithms. For CTR estimation, many fancy models like DeepFM,xDeepFM,DIN,DIEN~\cite{DeepFM,xDeepFM,DIN,DIEN} etc. have been proposed in the last few years. By taking advantage of great nonlinear fitting capability of deep neural network and huge amount of click and impression data, CTR models have achieved great performance. However, due to label collection and dataset size problems, it becomes quite different and challenging for CVR modeling.
	
	The major difficulties of CVR modeling can be summarized into the following three points. The first challenge is data sparsity. We usually train CVR models using dataset composed of clicked impressions, which is often quite small compared with CTR dataset. What's  even worse is that conversions are often extremely scarce. Data sparsity problem makes CVR model quite hard to fit and usually can not get satisfactory results as we expected. The second challenge is sample selection bias problem, which refers to the inconsistency of data distribution between training and testing space. Conventional CVR models are trained with clicked impressions but
	are used to make inference on the entire space with all impressions. This may hurt the generalization performance of CVR models a lot. The last but not least challenge is conversion delay problem. Different from click feedback can be collected immediately after the click event occurred, many conversions can only be observed after a relatively long and random delay since clicks happened, resulting in many false negative labels in the CVR dataset. For example, users may add products into carts or wishlist first. After several comparisons, he or she then decides to pay or not. This kind of delayed feedback creates lots of false negative samples and leads to underestimation of CVR modeling.
	
	To alleviate the problem of data sparsity,~\cite{HierarchicalEstimator} models the conversion event at different select hierarchical levels with separate binomial distributions and combines individual estimators using logistic regression. However, this method can not be applied in search and recommendation systems which has thousands of millions of users and items. ~\cite{AttCVR_DelayedFeedback} chooses to use pre-trained image models to generate dense item embeddings, which is greatly influenced by the quality of images. For solving sample selection bias problems, ~\cite{AMAN} introduces some negative examples by random sampling unclicked impressions, however, their method usually results in underestimated prediction.  \cite{ESMM,ESMM2} make use of the information of users' sequential behavior graph, for example, "impression $\rightarrow$ click $\rightarrow$ purchase" in  ESMM and more generally, "impression $\rightarrow$ click $\rightarrow$ D(O)Action $\rightarrow$ purchase" in  ESM$^2$. Though ESMM and ESM$^2$ greatly relieve the problems of data sparsity and sample selection bias, they both ignore the issue of conversion delay, which is a unique challenge of CVR modeling. To handle the time delay issue, ~\cite{Chapelle_DFM} assumes that the time delay follows an exponential distribution and introduces an additional model to capture conversion delay.  However, there is no guarantee that the time delay follows exponential distribution. \cite{Nonpara_DFM} proposes a more general non-parametric delayed feedback model to estimate the time delay without parametric distribution assumption, but the calculation of kernel density function is too complex to deploy in the industrial environment.
	
	In this paper, we propose an end to end framework from the perspective of entire space, trying to solve above three challenges at the same time.  By utilizing the sequential pattern of user actions, we employ a multi-task framework and construct two auxiliary tasks of CTR and CTCVR to eliminate the problem of data sparsity and sample selection bias. Besides, to tackle the time delay challenge, we design a novel time delay model. Without any special assumption about the time delay distribution, we approximate the procedure with the mechanism of survival analysis. Concretely, we transform the delayed time into discrete day slots, and approximate the survival probability by predicting which slot a conversion will fall into. The object optimization is conducted by maximizing log-likelihood of the dataset. The main contributions of this paper are summarized as follows:
	
	\begin{itemize}
		\item
		We combine the advantage of user sequential behavior pattern and the time delay factor from a perspective of entire space, and propose a novel ESDF framework which tackles the data sparsity, sample selection bias and time delay challenges of CVR prediction simultaneously.
		\item
		Without any special distribution assumption, ESDF transforms the delayed time into discrete slots and approximates the delay procedure with the mechanism of survival analysis, which is more general and practical.
		\item
		Extensive experiments on industrial datasets are conducted to show the effectiveness of ESDF. To the best of our knowledge, it is the first attempt to unitedly solve the above three challenges. Besides, we release a sample dataset for the future study, which is the first public dataset consisting of user sequential behaviors and time delay information on the entire space. The dataset can be found at the following url: $URL$\footnote{The address will be present in the camera-ready version}.
	\end{itemize}

	\begin{table*}[]
		\caption{Notations and descriptions}
		\centering
		\begin{tabular}{l|l}
			\hline
			Variable & Description \\
			\hline
			$X \in R^M $ & \tabincell{l}{A set of feature vectors representing samples of users and items, \\ where M is the dimension of the feature vector.} \\
			$Y \in \{0,1\}$ & Indicating whether a click has happened. \\
			$Z \in \{0,1\}$ & Indicating whether a conversion has already happened. \\
			$C \in \{0,1\}$ & Indicating whether a conversion will eventually happen or not. \\
			$D \in [0, \infty)$ & Time delay between the click and the conversion (undefined if $C = 0$). \\
			$E \in [0, \infty)$ & The elapsed time since the click happened. \\
			$S(e) = P(D \ge e)$ & The survival function. \\
			\hline
			$I_{1,1} = \{i|z_i =1 \& y_i =1, i=1,2...,N\}$ & A set of sample indices that are clicked and already converted. \\
			$I_{0,1} = \{i|z_i =0 \& y_i =1, i=1,2...,N\}$ & A set of sample indices that are clicked and not converted yet. \\
			$I_{0,0} = \{i|z_i =0 \& y_i =0, i=1,2...,N\}$ & A set of sample indices that are not clicked and not converted now. \\
			\hline
			$x_i, y_i, z_i, c_i, d_i, e_i$ & Values of the $i$th sample corresponding to the above variables. \\
			\hline
		\end{tabular}
		\label{table:Notation}
	\end{table*}
	
	\section{Related Work}
	In the last decade, CVR prediction models have received much attention and achieved remarkable effectiveness. Many well designed CVR models have been proposed and brought huge commercial revenues for companies and advertisers. ~\cite{HierarchicalEstimator,Chapelle_DFM} used logistic regression for modeling CVR problems. ~\cite{GBDT_Lu} utilized Gradient Boosted Decision Tree(GBDT) models to identify strong feature interactions and then constructed the data-driven trees. However, these models can not capture the underlying nonlinear relationships well and need lots of experts' hand-crafted features. With the development of deep learning and successful applications in different areas, like computer vision~\cite{ResidualNetwork}, natural language processing ~\cite{Bert,TransformerXL}, advertising CTR prediction\cite{DeepFM,DIN,DIEN}, etc., many studies attempt to take advantage of the strong representation ability of neural network to tackle the CVR problem. However, most of existing methods focus on one or two challenges mentioned above.
	
	For relieving the problem of data sparsity, ~\cite{Oversampling} copies the rare class samples. However, the algorithm is sensitive to sampling rates and not easy to achieve the consistently optimal result. ~\cite{HierarchicalEstimator} utilizes past performance observations along user, publisher and advertiser data hierarchies and models the conversion event at different select hierarchical levels with separate binomial distributions, then they use logistic regression to combine these individual estimators for final conversion events prediction. Nevertheless, it is hard to be applied in search and recommendation systems with hundreds of millions of users and items. ~\cite{AttCVR_DelayedFeedback} chooses to use pre-trained image models to generate dense item embeddings and substitute sparse ID embeddings. This helps relieve the data sparsity problem but is greatly influenced by the quality of images.
	
	The second largest challenge is sample selection bias problem.  CVR dataset is usually collected from clicked samples.  It becomes different at inference stage where the prediction is over the entire space.  Different distribution between training and inference stage greatly limits the generalization of CVR model. In order to solve such problem, ~\cite{AMAN} proposes All Missing As Negative(AMAN) and selects unclicked samples as negative examples. However, their method usually results in underestimated predictions and the proportion of negative examples is hard to decide. ~\cite{ESMM} transforms the CVR problem into a multi-task problem. By making use of the user sequential actions, "impression $\rightarrow$ click  $\rightarrow$  pay", ESMM models CVR directly over the entire space. This greatly reduces the problem of sample selection bias and data sparsity. ~\cite{ESMM2} extends the user sequential actions to a more general situation, "impression $\rightarrow$ click  $\rightarrow$ D(O)Action  $\rightarrow$  pay". ESMM and ESM$^2$ both relieve the problem of data sparsity and sample selection bias. However, they all ignore the third unique problem in CVR modeling: time delay problem.
	
	Time delay problem is quite important in CVR predictions. As shown by \cite{Chapelle_DFM}, more than 50\% conversion happens after 24 hours, which brings many false negative samples and misleads the model training. To relieve this problem, \cite{Chapelle_DFM} assumes that the conversion delay follows an exponential distribution, \cite{Ji_Weibull} thinks the distribution is a mixture of Weibull distribution. Nevertheless, the actual distribution of time delays is not guaranteed to be exponential nor Weibull. Recently, ~\cite{Nonpara_DFM} proposes a more general non-parametric delayed feedback model and models the distribution of time delay depending on the content of an ad and the features of a user. Though these methods help relieve the problem of time delay, they ignore the sample selection bias problem and do not apply deep neural network technology, which makes these methods difficult to capture the underline nonlinear relationships. To the best of our knowledge, there not exists previous works that take three problems into account at the same time.

	\section{Algorithm}
\begin{figure}[htbp]
		\centering
		\includegraphics[width=1.0\columnwidth]{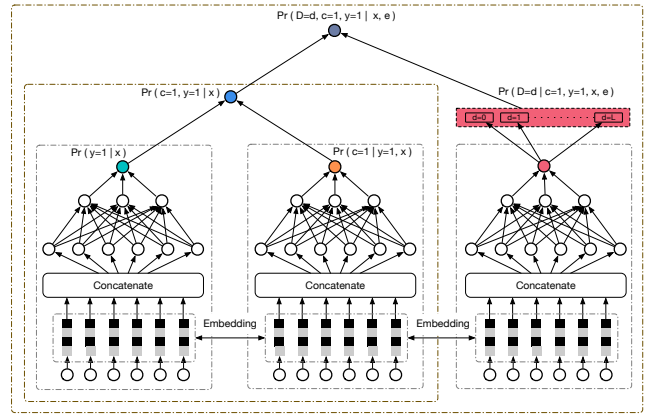}
		\caption{Model Framework}.
		\label{fig:ESDF}
	\end{figure}

	In this section, we will introduce our neural network framework ESDF, which consists of two main parts: conversion model part and time delay model part.  The whole neural network framework is shown in figure  \ref{fig:ESDF}. For better explanation, we summarize the notations of variables used for ESDF in Table \ref{table:Notation}. In the following parts, we will split our model into three modules. At the first part, we will explain the details of our conversion model and how we deal with data sparsity and sample selection bias problems. At the second part, we will describe the details about how we construct time delay model and the difference between other time delay models. By combining the conversion and time delay models together, we introduce our approach in the final part.

	\subsection{Conversion Model}
	The target of conversion model is to estimate the probability of $pCVR = P(C =1 | Y = 1,X=x,E=e)$.  Similar to many previous work like \cite{Chapelle_DFM,AttCVR_DelayedFeedback}, we assume that:
	
	\begin {equation}
	\begin{aligned}
		P(C|Y = 1,X,E) = P(C | Y=1,X) \label{assumption}
	\end{aligned}
	\end {equation}
	
	This makes sense because the elapsed time since the click usually does not make any influence on the final conversion. To eliminate the sample selection bias and data sparsity problem, we refer to the idea of ESMM and ESM$^2$ ~\cite{ESMM,ESMM2}. By making use of the sequential pattern of user actions "impression $\rightarrow$ click $\rightarrow$ pay", we construct CTR and post-view click-through \& conversion rate (CTCVR) tasks together. Here, we use $pCTR$ and $pCTCVR$ to represent the probability of CTR and CTCVR respectively, where $pCTR=P(Y=1|X=x)$ and $pCTCVR=P(C=1,Y=1|X=x)$. The CVR prediction of the $i$th sample can then be formulated as:
	
	\begin {equation}
	P(c_i=1 | y_i=1,x_i) = \frac{P(c_i=1,y_i=1|x_i)}{P(y_i=1|x_i)} \label{eq1}
	\end {equation}
	
	As can be seen from the equation \ref{eq1},  $P(c_i=1,y_i=1|x_i)$ and $P(y_i=1|x_i)$ can all be modeled on the entire input space $X$, which greatly helps relieve the problem of sample selection bias. In most situations, the click rate is small, so $P(c_i=1 | y_i=1,x_i) $ may be larger than 1 and may cause numerical instability. Hence, we model $pCTR$ and $pCTCVR$ at the same time. This helps us avoid the problem of numerical instability. Besides, we are able to model with all CTR samples which helps us get more stable training process and better performance, because the size of CTR samples is much larger than the size of CVR samples.  In our neural network framework, the embedding layer is shared which enables CVR network to learn from huge volume of CTR samples and alleviate the data sparsity trouble a lot. Different from ~\cite{ESMM,ESMM2}, we add time delay model attempting to solve the problem of feedback delay problem.
	
	\subsection{Time Delay Model}
	
	To address the problem of feedback delay,  we model the time delay with reference to the idea of survival analysis\cite{SurvivalAnalysis}. Survival analysis is originally used to analyze the expected duration of time until one or more events happen. Corresponding to our problem, conversion can be considered as an "event" here.  We use $f(t)$ to denote the probability density function, which represents the probability that the  conversion will happen at time $t$. Then we can get our survival function:
	\begin {equation}
	\begin{aligned}
		S(t) = 1 - \int_{0}^{t}f(t)dt
	\end{aligned}
	\end {equation}
	
	To make our model more general and suitable for different scenarios. We do not make any special assumptions of the distribution, like exponential distribution \cite{Chapelle_DFM} or Weibull distribution ~\cite{Ji_Weibull}. Instead, we choose to make use of the features of users, items and context to construct the time delay model, which is more reasonable and easy to generalize to other scenarios. In our algorithm, we first transform the delayed time via day slots. Concretely, the delayed time are divided into $T+2$ bins. The $i$th bin where $i \le T$ represents that the conversion happens at $i$ days later after the click and $T+1$ indicates that the delay time of conversion is bigger than $T$ days. This is quite reasonable because when $T$ is large, the conversions only occupy a small part and can be recognized as noise. Hence, we put all conversions happened after $T$ days later since the click into the $T+1$ bin. The probability that the conversion happens at the $t$ days since the click can be described as:
	\begin {equation}
	\begin{aligned}
		P(D=t|C=1,Y=1,E=e,x) = F(g(x, e),t)
	\end{aligned}
	\end {equation}
	where $g(x, e)$ represents the softmax output of the time delay model and $F(\cdot,t)$ is the $t$th value of $g(x, e)$, $t \in [0,T+1]$. For simplicity, we will use $f(t,x,e)$ to represent $F(g(x, e),t)$ in the following content.
	
	
	\subsection{Proposed Approach}
	By combining the two models of conversion and time delay, we can get the final ESDF. Concrete relations within ESDF can be summarized as below:
	
	
	\begin{itemize}
		\item
		If a conversion has already happened, we have $Z = 1$, then $C$ must be 1 and the delay time $D$ can be observed.
		\begin{equation}
		Z = 1 \rightarrow C = 1 ~ and ~ E > D = d \label{relation1}
		\end{equation}
		\item
		If a conversion has not happened, we have $Z = 0$. It is either because the user will not convert or because he will convert later, in other words,
		\begin{equation}
		Z = 0 \rightarrow C = 0 ~ or ~ E < D \label{relation2}
		\end{equation}
	\end{itemize}
	For the sake of convenience, we define three sets of sample indices:
	\begin {equation}
	\begin{aligned}
		I_{1,1} & = \{i|z_i =1 \& y_i =1, i=1,2...,N\} \\
		I_{0,1} & = \{i|z_i =0 \& y_i =1, i=1,2...,N\} \\
		I_{0,0} & = \{i|z_i =0 \& y_i =0, i=1,2...,N\} \\
	\end{aligned}
	\end {equation}
	We respectively refer to $I_{1,1}, I_{0,1}, I_{0,0}$ as the observed conversions, unobserved conversion and unclicked samples.
	Given parameters $\Theta = \{\theta_{ctr}, \theta_{ctcvr}, \theta_{delay}\}$, the likelihood of ESDF $P(\mathcal{D};\Theta)$ can be factorized as:
	
	\begin {equation}
	\begin{aligned}
		P(\mathcal{D};\Theta) & = \mathop{\Pi}\limits_{i \in I_{1,1}} P(z_i=1, y_i = 1 |x_i, e_i) \\
		& \times \mathop{\Pi}\limits_{i \in I_{0,1}} P(z_i=0, y_i = 1 |x_i, e_i) \\
		& \times \mathop{\Pi}\limits_{i \in I_{0,0}} P(z_i=0, y_i = 0 |x_i, e_i) \\
	\end{aligned}
	\end {equation}
	Under the assumption (\ref{assumption}) and the equivalence (\ref{relation1}), the probability $P(z_i=1, y_i = 1 |x_i, e_i)$ can by written as:
	
	\begin {equation}
	\begin{aligned}
		&P(z_i=1, y_i = 1 | x_i, e_i)  \\
		=&P(c_i=1, D=d_i, y_i = 1 | x_i,e_i)  \\
		=&P(c_i=1, y_i=1 | x_i, e_i) P(D=d_i|c_i=1,y_i=1,x_i,e_i)  \\
		=&P(c_i=1, y_i=1 | x_i) P(D=d_i|c_i=1,y_i=1,x_i,e_i)  \\
		=&q_i f(d_i, x_i, e_i) \label{dev1}
	\end{aligned}
	\end {equation}
	where $q_i = P(c_i=1,y_i=1|x_i)$ and $f(d_i, x_i, e_i) = P(D=d_i|c_i=1,y_i=1,x_i,e_i)$ in the last equation. The first equation comes from the relation (\ref{relation1}), while  the third equation comes from the conditional independence (\ref{assumption}). Note that when a conversion has been observed, $e_i$ has to be greater or equal to $d_i$ here. Furthermore, the probability that sample $i$ has been clicked and the conversion has not been observed yet can be formulated as:
	
	\begin {equation}
	\begin{aligned}
		&P(z_i=0, y_i=1 | x_i, e_i)  \\
		=&P(z_i=0, c_i=0, y_i=1 | x_i,e_i) \\
		+&P(z_i=0, c_i=1, y_i=1 | x_i,e_i) \\
		=&P(c_i=0, y_i=1 | x_i) \\
		+&P(z_i=0, c_i=1, y_i=1 | x_i,e_i) \\
		=&P(y_i=1 | x_i) - P(c_i=1, y_i=1 | x_i) \\
		+&P(c_i=1, y_i=1 | x_i) P(z_i=0 | c_i=1, y_i=1, x_i,e_i) \\
		=&P(y_i=1 | x_i) - q_i + q_i P(z_i=0|c_i=1,y_i=1, x_i,e_i)  \label{dev2}
	\end{aligned}
	\end {equation}
	The second equation comes from the conditional independence (\ref{assumption}) and the truth $P(z_i = 0|c_i=0,y_i=1,x_i,e_i) = 1$. The probability of delayed conversion is:
	\begin {equation}
	\begin{aligned}
		&P(z_i=0 | c_i=1, y_i=1, x_i, e_i) \\
		=&P(D>e_i | c_i=1, y_i=1, x_i, e_i) \\
		=&\sum_{t = e_i + 1}^{T+1} f(t,x_i,e_i)  \label{dev3}
	\end{aligned}
	\end {equation}
	Finally, the probability that sample $i$ is not clicked is:
	\begin {equation}
	\begin{aligned}
		&P(z_i=0,y_i=0|x_i, e_i)  \\
		=&P(y_i=0|x_i) P(z_i=0|y_i=0,x_i) \\
		=&1 - P(y_i=1|x_i) \label{dev4}
	\end{aligned}
	\end {equation}
	The second equation comes from the truth: $P(z_i=0 | y_i=0, x_i) = 1$. By substituting the formula (\ref{dev1}),(\ref{dev2}), (\ref{dev3}),(\ref{dev4}) into the likelihood function, we can get the following result. Note that for simplicity, we use $p_i$ to represent $P(y_i=1|x_i)$
	
	\begin {equation}
	\begin{aligned}
		P(\mathcal{D};\Theta) & = \mathop{\Pi}\limits_{i \in I_{1,1}} q_i f(d_i,x_i,e_i) \\
		& \times \mathop{\Pi}\limits_{i \in I_{0,1}} [p_i - q_i + q_i \sum_{t = e_i + 1}^{T+1} f(t,x_i,e_i)] \\
		& \times \mathop{\Pi}\limits_{i \in I_{0,0}} (1 - p_i) \label{dev5}
	\end{aligned}
	\end {equation}
	
	\subsection{Learning Algorithm}
	In this subsection, we explain the learning algorithm for ESDF derived on the basis of Expectation-Maximization (EM) algorithm. EM algorithm is widely used to optimize the maximum likelihood estimation of probabilistic models that depend on hidden variables. Due to the time delay issue, the variable $C$ can not be observed immediately after click occurred. Hence, we treat variable $C$ as the hidden variable. Specifically, we calculate the expectation of $C$ with current model in the expectation step, and maximize the expected log-likelihood according to the expectation of $C$ in the maximization step.
	
	$\textbf{Expectation Step}$ For a given sample, we need to calculate the posterior probability of the hidden variable. We denote the expectation for sample $i$ as $w_i$. It is trivial to get the following equations:
	
	\begin {equation}
	\begin{aligned}
		w_i &= 1, \forall i \in I_{1,1} \\
		w_i &= 0, \forall i \in I_{0,0} \label{dev6}
	\end{aligned}
	\end {equation}
	The first equation comes from the relation (\ref{relation1}). For $i \in I_{0,1}$, we can not observe the conversion directly and need to estimate the expectation of $C$ given $Z = 0 ~\&~ Y = 1$.
	
	\begin {equation}
	\begin{aligned}
		w_i &= P(c_i=1 | z_i = 0, y_i = 1, x_i, e_i) \\
		&= \frac{P(c_i=1 | y_i=1, x_i) P(z_i=0 | c_i=1, y_i=1, x_i, e_i)}{P(z_i=0 | y_i=1, x_i, e_i)} \\
		&=\frac{q_i\sum_{t = e_i + 1}^{T+1} f(t,x_i,e_i)}{p_i - q_i + q_i \sum_{t = e_i + 1}^{T+1} f(t,x_i,e_i)} \label{dev7}
	\end{aligned}
	\end {equation}
	
	$\textbf{Maximization Step}$ In maximization step, we maximize the expected log-likelihood according to the distribution computed during the expectation step. Based on equations of (\ref{dev5}) and (\ref{dev6}), the expected log-likelihood loss for $I_{1,1}$ and $I_{0,0}$ can be formulated as:
	
	\begin {equation}
	\begin{aligned}
		\mathcal{L}_{1,1} &= \sum_{i \in I_{1,1}} {w_i[~log(q_i) + log(f(d_i,x_i,e_i))~]} \\
		&= \sum_{i \in I_{1,1}} {[~log(q_i) + log(f(d_i,x_i,e_i))~]} \\
		\mathcal{L}_{0,0} &= \sum_{u \in I_{0,0}} {(1-w_i)log(1 - p_i)} \\
		&= \sum_{u \in I_{0,0}} {log(1 - p_i)} \label{loss1}
	\end{aligned}
	\end {equation}
	For the expected log-likelihood of $I_{0,1}$, we split it into two parts with $C=0$ and $C=1$, corresponding to the $p_i - q_i$ and $q_i\sum_{t = e_i + 1}^{T+1}{f(t,x_i,e_i)}$ of (\ref{dev5}) respectively. The expected log-likelihood can be formulated as:
	\begin {equation}
	\begin{aligned}
		\mathcal{L}_{0,1} &= \sum_{i \in I_{0,1}} {w_i[~log(q_i) + log(\sum_{t = e_i + 1}^{T+1}{f(t,x_i,e_i)})~]} \\
		&+ \sum_{i \in I_{0,1}} {(1-w_i)~log(p_i - q_i)} \label{loss2}
	\end{aligned}
	\end {equation}
	Combining the equations of (\ref{loss1}) and (\ref{loss2}), the quantity to be maximized during maximization step can finally be summarized as:
	
	\begin {equation}
	\begin{aligned}
		&\mathcal{L}(\mathcal{D};\Theta) = \sum_{i}{w_i log(q_i)} \\
		&+\sum_{i}{(1-w_i)[y_i log(p_i-q_i) + (1-y_i)log(1-p_i)]} \\
		&+\sum_{i}{w_i z_i log(f(t_i,x_i,e_i))} \\
		&+\sum_{i}{w_i(1-z_i)log(\sum_{t = t_i + 1}^{T+1}{f(t,x_i,e_i)})} \\
	\end{aligned}
	\end {equation}
	with
	
	\begin{equation}
	\begin{aligned}
	t_i = \left\{
	\begin{array}{rcl}
	e_i & if & z_i = 0 \\
	d_i & if & z_i = 1 \\
	\end{array}
	\right.
	\end{aligned}
	\end {equation}
	
	\section{Experiments}
	
	In this section, we will present our empirical studies of the proposed method. The main purpose of experiments is to illustrate two characters of our algorithm.
	
	\begin{itemize}
	\item
	Effectiveness: our proposed model ESDF tackles data sparsity, sample selection bias and time delay problems at the same time. To verify the effectiveness of our model, we conduct extensive experiments to show that ESDF can achieve the comparable performance among all comparison methods.
	\item
	Superiority of time delay model: By discretizing the delay time and taking advantage of the representation capability of deep neural network, ESDF is a more general and practical solution for industrial situations. We show the superiority of time delay by comparing with different kinds of time delay processing methods.
	\end{itemize}
	
	\subsection{Datasets}
	
\begin{figure}[t]
	\centering
	\includegraphics[width=1.0\columnwidth]{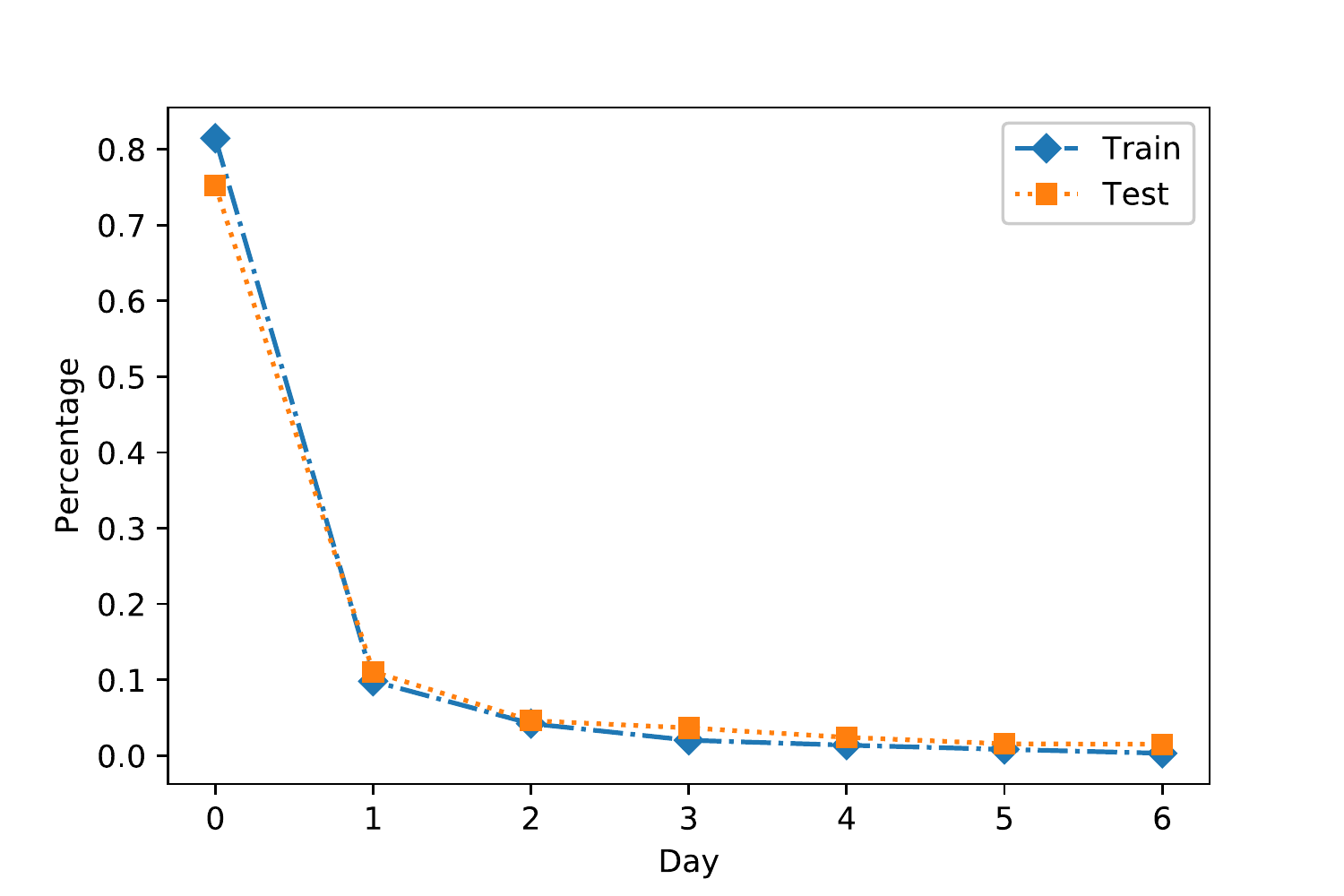}
	\caption{Delayed conversion distribution}.
	\label{fig:conversion_distribution}
	\end{figure}

	According to our survey, it's the first time to model the CVR prediction with delayed feedback on the entire space. Most of the related datasets are collected during the post-click stage like the classical criteo dataset used in ~\cite{Chapelle_DFM}, or lack of the delayed information as ~\cite{ESMM}. No public datasets consist of impression, click and conversion labels simultaneously in CVR modeling area. To evaluate the proposed method and better study this problem in the future, we collect traffic logs from our E-commerce search system and opensource a simplified random sampling version of the whole dataset. In the rest of this paper, we refer to the released dataset as Public Dataset and the whole dataset as Product Dataset. The statistics of the two datasets are summarized in Table \ref{table:dataset}. The training data is made of logs from 2020-05-30 to 2020-06-05, while the testing data consists of data on 2020-06-06. For the delayed conversion samples, we set the attribution window to 7 days. So the conversion labels of testing data have been attributed up to 2020-06-12, while labels of the training data can only be attributed up to 2020-06-05. This is consistent with the actual production environment since we can not observe the future information. As shown in figure \ref{fig:conversion_distribution}, about 80$\%$ conversions occur on the first day, but the rest of them occur much later. It means that we may mistake 20\% positive samples for negative if we adopt the traditional label collection paradigm within one day. There exits a slight difference of the distribution between training and testing data. It is because the testing data is labeled with the ground truth conversion. More detailed descriptions of the dataset can be found in the website of $URL$\footnotemark[1].
	
	\begin{table*}[t]
	\caption{Summary of the data structure}
	\centering
	\label{table:dataset}
	\begin{tabular}{lcccccc}
	\hline
	Dataset         & Feature Dimension & \# Feature Field& \# Impression & \# Click & \# Conversion & Attribution Window \\
	\hline
	Public Dataset  & 1.7B & 240 & 30.6M & 0.74M & 14.7K & 7 days \\
	Product Dataset & 5.0B & 544 & 11.1B & 291M & 5.53M & 7 days \\
	\hline
	\end{tabular}
	\end{table*}
	
	\subsection{Experimental Settings}
	
	$\textbf{Metric}$ ROC AUC ~\cite{AUC} is a widely used metric in CVR prediction task, it denotes the probability of ranking a random positive sample higher than a negative sample. A variation of group AUC (GAUC) is introduced in \cite{GAUC1,GAUC2} which measures the items of intra-user order by averaging AUC over users and is shown to be more relevant to online performance on CVR prediction. In this paper, we group the impressions by each request of users to compute the GAUC:
	
	$$GAUC = \frac{\sum_{i=1}^{N}{\#impression_{i} \times AUC_{i}}}{\sum_{i}^{N} \#impression_{i}}$$
	
	where $N$ is the number of user requests and $AUC_{i}$ corresponds to the performance of ranking model for the $i$th request. It's worth noting that the value of GAUC is 0.5 for a group of samples with the same label. When computing GAUC, we just only consider groups comprising both positive and negative samples in practice. Besides, we follow ~\cite{RelaImpr} to introduce RelaImpr to measure the relative improvement over models. The RelaImpr is defined as below:
	$$RelaImpr = (\frac{AUC(measured model) - 0.5}{AUC(base model) - 0.5} - 1) \times 100\%$$
	
	$\textbf{Compared Methods}$ To verify the effectiveness of our delayed feedback framework, we choose the following baselines for comparison. To assure fairness, we re-implement all methods with deep neural work. All baselines share embeddings between CTR and CVR tasks.
	\begin{itemize}
	\item
	ESMM ~\cite{ESMM}: ESMM eliminates the data sparsity and selection bias problems by modeling CVR on the entire space. Here we use it as the baseline framework. Following the baseline method in ~\cite{Chapelle_DFM}, samples for which the conversion is unobserved in the first day are treated as negative. It is a traditional paradigm to process industrial data in practice.
	\item
	NAIVE: To evaluate the impact of false negative samples, we remove them from training set and build a naive competitor with the same model of ESMM.
	\item
	SHIFT: SHIFT extends the attribution window from one day to seven days. False negative samples are gradually corrected day by day limited to the newest date we can observe. This method partially fixes the false negative labels but doesn't consider the elapsed time and delay distribution.
	\item
	DFM~\cite{Chapelle_DFM}: As a strong competitor, DFM considers the delayed feedback with an exponential distribution assumption for the time delay. A lot of methods based on hazard function approximation finally degenerate into DFM after assuming that the hazard function is independent with the elapsed time. For a fairness compare, we replace the logistic regression of DFM with deep neural network and also add the embedding share between CTR and CVR network. The new version of DFM has the same network backbone with our method.
	\end{itemize}

	$\textbf{Parameter Settings}$ For all experiments, the dimension of embedding is set to 8 and the batch size is 1024. We use Adam ~\cite{Adam} solver with an initial learning rate of 0.0001. Both CTR and CVR branches share the same MLP architecture of $512 \times 256 \times 128 \times 1$ for all models. ReLU is used to be the activation function. The number of slots for the time delay discretization is set to 7. We report all results on public and product datasets with the same hyperparameters.

	\subsection{Effectiveness of ESDF}
	
	Experimental results on public and product datasets are reported in Table \ref{exp_results}. Remind that, for the public dataset, impressions from the same request are sparse due to the random sampling, and GAUC by individual request will collapse to 0.5. So we adopt the ROC AUC as the metric on public dataset and GAUC on product dataset.  As can be seen from Table \ref{exp_results}, ESDF outperforms the other competitors both in public and product datasets and achieves almost 0.08 absolute GAUC gain and 6.68\% RelaImpr over the ESMM baseline, which is considered substantial for our application.
	
	\begin{table}[t]
	\caption{Comparison of different models.}
	\centering
	\begin{tabular}{lcccc}
	\hline
	\multirow{2}{*}{Model}  & \multicolumn{2}{c}{Public Dataset} & \multicolumn{2}{c}{Product Dataset} \\
	& AUC       & RelaImpr       & GAUC       & RelaImpr       \\
	\hline
	\multicolumn{1}{l}{ESMM}        & 0.7679    & 0.00\%           & 0.6107    & 0.00\%   \\
	\multicolumn{1}{l}{NAIVE}        & 0.7693    & 0.52\%        & 0.6112    & 0.45\%   \\
	\multicolumn{1}{l}{SHIFT}        & 0.7740    & 2.28\%           & 0.6136    & 2.62\%   \\
	\multicolumn{1}{l}{DFM}            & 0.7789    & 4.11\%        & 0.6146    & 3.52\%   \\
	\multicolumn{1}{l}{\textbf{ESDF}} & \textbf{0.7811}    & \textbf{4.93\%}    & \textbf{0.6181}    & \textbf{6.68\%}       \\
	\hline
	\end{tabular}
	\label{exp_results}
	\end{table}
	
	\subsection{Superiority of Time Delay Model}
	We can observe that the baseline method ESMM achieves the worst performance because it contains many false negative samples. By simply removing these noise samples, NAIVE gets a better result, around 0.52\% and 0.45\% improvement of RelaImpr on the public and product datasets respectively. SHIFT makes a further improvement by partially correcting the false negative samples and achieves 2.28 \% and 2.62\% improvements compared with simple ESMM. However, due to the lack of consideration of the feedback delay distribution, SHIFT can not compete against DFM. As a strong competitor, DFM makes a significant improvement compared with above three methods. There are two main factors that restrict DFM after our re-implementation with deep neural network: First, it ignores the user sequential behavior pattern which can relieve the impact of sample selection bias; Second, the exponential distribution assumption limits the hypothesis space of feedback delay model. ESDF makes an improvement from these two aspects and achieves 0.82\% and 3.16\% improvement of RelaImpr over DFM on the public and product datasets respectively.
	
	
	\subsection{Analysis of the Delayed Feedback Samples}
	
	\begin{figure}[t]
	\centering
	\subfloat[Public Dataset]{
		\label{fig:sub_a}
		\begin{minipage}[t]{0.49\columnwidth}
		\centering
		\includegraphics[width=1.0\columnwidth]{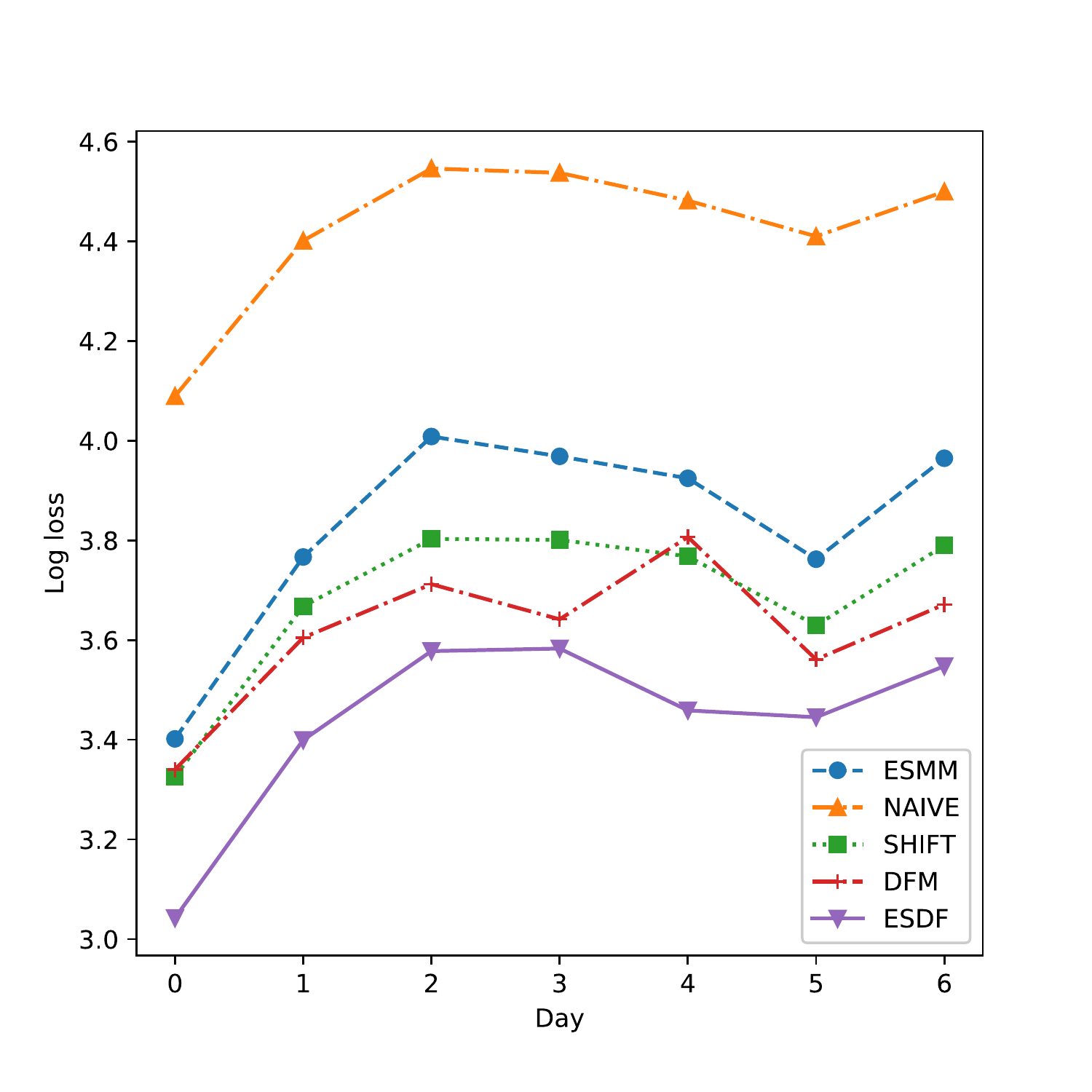}
		\end{minipage}
	}
	\subfloat[Product Dataset]{
		\label{fig:sub_b}
		\begin{minipage}[t]{0.49\columnwidth}
		\centering
		\includegraphics[width=1.0\columnwidth]{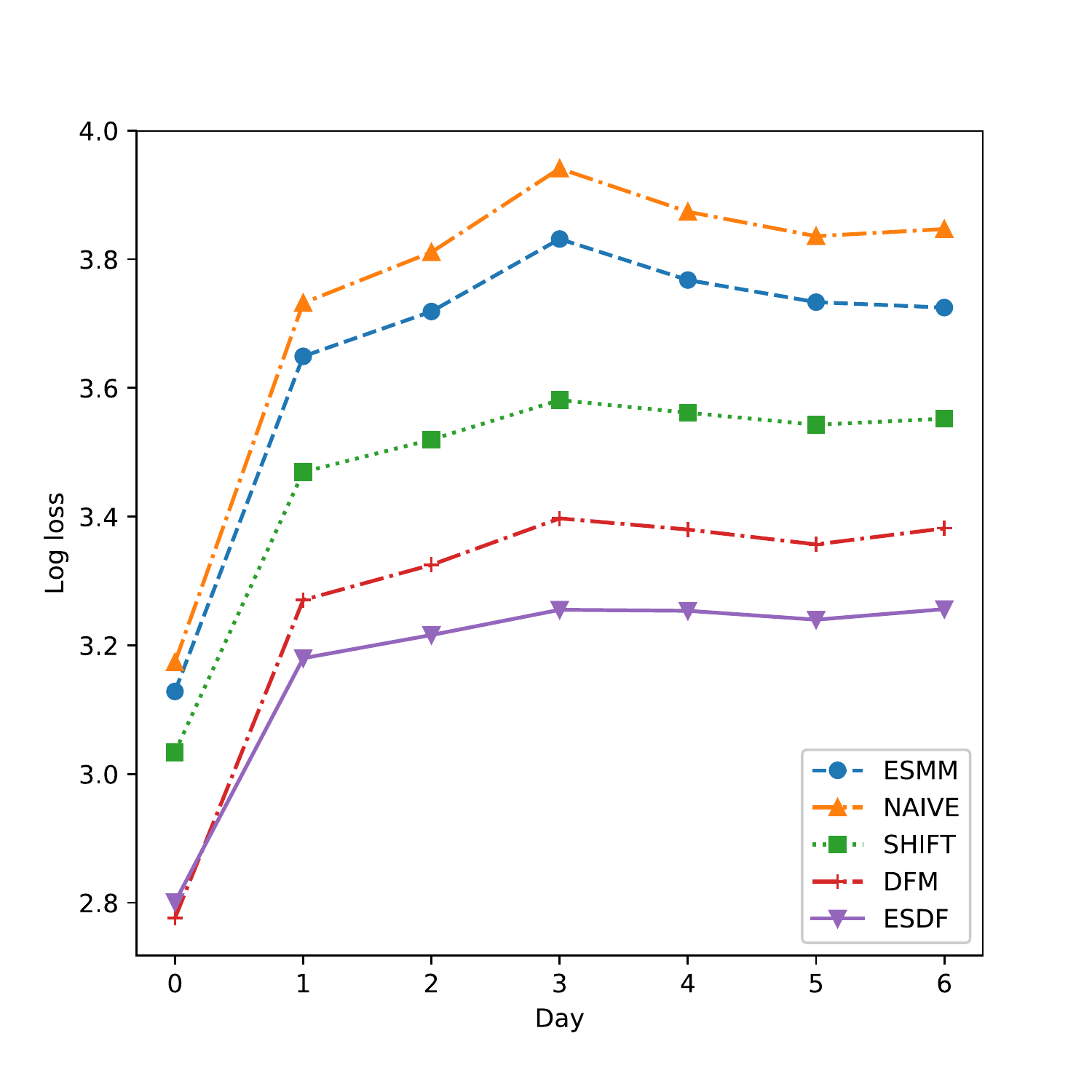}
		\end{minipage}
	}
	\caption{Log loss of delayed feedback samples}
	\label{fig:log_loss}
	\end{figure}
	
	In this subsection, we focus on the performance of different models on the delayed samples. For comparison, we adopt log loss as the metric which is more attentive to the precision of models. We try to get an insight into the impact of delayed feedback samples to different methods. As shown in Figure \ref{fig:log_loss}, all models, as expected, perform better on data of the first day. The feedback delay hides the original distribution and makes the prediction imprecise. Time delay methods of DFM and ESDF outperform the others on both public and product datasets, but DFM is a bit unstable on the public dataset. We believe it is related to the inconsistency between true distribution of data and the assuming exponential one. An interesting fact is that, unlike the observations of GAUC, the naive method performs worse than ESMM, especially on the public dataset. A possible reason is that removing delayed feedback samples increases the difference between positive and negative classes, reducing the difficulty of model learning around the classification boundary. But on the other side, it expands the gap between training and testing datasets, making the model tend to give an imprecise prediction on the unseen data.

	\section{Conclusions}
	In this paper, we propose a novel neural network framework to unitedly tackle data sparsity, sample selection bias and feedback delay challenges in CVR prediction. Unlike existing methods, ESDF models the CVR prediction from a perspective of entire space, combining the advantage of user sequential behavior pattern and the time delay factor. Without special distribution assumptions, ESDF discretizes the delay time by day slot and models the probability based on survival analysis, which avoids the complex integral calculation and is friendly to the real application. Extensive experiments are conducted to evaluate the effectiveness of our method, and a public dataset is released for future study. More systematic methods to tackle these challenges from a unified perspective need to be investigated.
	
	\clearpage
	\bibliography{ref}

\begin{thebibliography}{23}
\providecommand{\natexlab}[1]{#1}
\providecommand{\url}[1]{\texttt{#1}}
\providecommand{\urlprefix}{URL }
\expandafter\ifx\csname urlstyle\endcsname\relax
  \providecommand{\doi}[1]{doi:\discretionary{}{}{}#1}\else
  \providecommand{\doi}{doi:\discretionary{}{}{}\begingroup
  \urlstyle{rm}\Url}\fi

\bibitem[{Chapelle(2014)}]{Chapelle_DFM}
Chapelle, O. 2014.
\newblock Modeling delayed feedback in display advertising.
\newblock In \emph{The 20th {ACM} {SIGKDD} International Conference on
  Knowledge Discovery and Data Mining}, 1097--1105.

\bibitem[{Dai et~al.(2019)Dai, Yang, Yang, Carbonell, Le, and
  Salakhutdinov}]{TransformerXL}
Dai, Z.; Yang, Z.; Yang, Y.; Carbonell, J.~G.; Le, Q.~V.; and Salakhutdinov, R.
  2019.
\newblock Transformer-XL: Attentive Language Models beyond a Fixed-Length
  Context.
\newblock In \emph{Proceedings of the 57th Conference of the Association for
  Computational Linguistics}, 2978--2988.

\bibitem[{Devlin et~al.(2019)Devlin, Chang, Lee, and Toutanova}]{Bert}
Devlin, J.; Chang, M.; Lee, K.; and Toutanova, K. 2019.
\newblock {BERT:} Pre-training of Deep Bidirectional Transformers for Language
  Understanding.
\newblock In \emph{Proceedings of the 2019 Conference of the North American
  Chapter of the Association for Computational Linguistics: Human Language
  Technologies}, 4171--4186.

\bibitem[{Fawcett(2006)}]{AUC}
Fawcett, T. 2006.
\newblock An introduction to ROC analysis.
\newblock \emph{Pattern recognition letters} 27(8): 861--874.

\bibitem[{Guo et~al.(2017)Guo, Tang, Ye, Li, and He}]{DeepFM}
Guo, H.; Tang, R.; Ye, Y.; Li, Z.; and He, X. 2017.
\newblock DeepFM: {A} Factorization-Machine based Neural Network for {CTR}
  Prediction.
\newblock In \emph{Proceedings of the 26th International Joint Conference on
  Artificial Intelligence}, 1725--1731.

\bibitem[{He et~al.(2016)He, Zhang, Ren, and Sun}]{ResidualNetwork}
He, K.; Zhang, X.; Ren, S.; and Sun, J. 2016.
\newblock Deep Residual Learning for Image Recognition.
\newblock In \emph{{IEEE} Conference on Computer Vision and Pattern
  Recognition}, 770--778.

\bibitem[{He and McAuley(2016)}]{GAUC1}
He, R.; and McAuley, J. 2016.
\newblock Ups and downs: Modeling the visual evolution of fashion trends with
  one-class collaborative filtering.
\newblock In \emph{proceedings of the 25th international conference on world
  wide web}, 507--517.

\bibitem[{Ji, Wang, and Zhu(2017)}]{Ji_Weibull}
Ji, W.; Wang, X.; and Zhu, F. 2017.
\newblock Time-aware conversion prediction.
\newblock \emph{Frontiers Comput. Sci.} 11(4): 702--716.

\bibitem[{Jr.(2011)}]{SurvivalAnalysis}
Jr., R. G.~M. 2011.
\newblock \emph{Survival analysis}, volume~66.

\bibitem[{Kingma and Ba(2014)}]{Adam}
Kingma, D.~P.; and Ba, J. 2014.
\newblock Adam: A method for stochastic optimization.
\newblock \emph{arXiv preprint arXiv:1412.6980} .

\bibitem[{Lee et~al.(2012)Lee, Orten, Dasdan, and Li}]{HierarchicalEstimator}
Lee, K.; Orten, B.; Dasdan, A.; and Li, W. 2012.
\newblock Estimating conversion rate in display advertising from past
  performance data.
\newblock In \emph{The 18th {ACM} {SIGKDD} International Conference on
  Knowledge Discovery and Data Mining}, 768--776.

\bibitem[{Lian et~al.(2018)Lian, Zhou, Zhang, Chen, Xie, and Sun}]{xDeepFM}
Lian, J.; Zhou, X.; Zhang, F.; Chen, Z.; Xie, X.; and Sun, G. 2018.
\newblock xDeepFM: Combining Explicit and Implicit Feature Interactions for
  Recommender Systems.
\newblock In \emph{Proceedings of the 24th {ACM} {SIGKDD} International
  Conference on Knowledge Discovery {\&} Data Mining}, 1754--1763.

\bibitem[{Lu et~al.(2017)Lu, Pan, Wang, Pan, Wan, and Yang}]{GBDT_Lu}
Lu, Q.; Pan, S.; Wang, L.; Pan, J.; Wan, F.; and Yang, H. 2017.
\newblock A Practical Framework of Conversion Rate Prediction for Online
  Display Advertising.
\newblock In \emph{Proceedings of the ADKDD}, 9:1--9:9.

\bibitem[{Ma et~al.(2018)Ma, Zhao, Huang, Wang, Hu, Zhu, and Gai}]{ESMM}
Ma, X.; Zhao, L.; Huang, G.; Wang, Z.; Hu, Z.; Zhu, X.; and Gai, K. 2018.
\newblock Entire Space Multi-Task Model: An Effective Approach for Estimating
  Post-Click Conversion Rate.
\newblock In \emph{The 41st International {ACM} {SIGIR} Conference on Research
  {\&} Development in Information Retrieval}, 1137--1140.

\bibitem[{Pan et~al.(2008)Pan, Zhou, Cao, Liu, Lukose, Scholz, and Yang}]{AMAN}
Pan, R.; Zhou, Y.; Cao, B.; Liu, N.~N.; Lukose, R.~M.; Scholz, M.; and Yang, Q.
  2008.
\newblock One-Class Collaborative Filtering.
\newblock In \emph{Proceedings of the 8th {IEEE} International Conference on
  Data Mining}, 502--511.

\bibitem[{Su et~al.(2020)Su, Zhang, Dai, Zhang, Yan, Wang, Bao, Xu, He, and
  Yan}]{AttCVR_DelayedFeedback}
Su, Y.; Zhang, L.; Dai, Q.; Zhang, B.; Yan, J.; Wang, D.; Bao, Y.; Xu, S.; He,
  Y.; and Yan, W. 2020.
\newblock An Attention-based Model for Conversion Rate Prediction with Delayed
  Feedback via Post-click Calibration.
\newblock In \emph{Proceedings of the 29th International Joint Conference on
  Artificial Intelligence}, 3522--3528.

\bibitem[{Weiss(2004)}]{Oversampling}
Weiss, G.~M. 2004.
\newblock Mining with rarity: a unifying framework.
\newblock \emph{Special Interest Group on Knowledge Discovery and Data Mining
  Explorations} 6(1): 7--19.

\bibitem[{Wen et~al.(2020)Wen, Zhang, Wang, Lv, Bao, Lin, and Yang}]{ESMM2}
Wen, H.; Zhang, J.; Wang, Y.; Lv, F.; Bao, W.; Lin, Q.; and Yang, K. 2020.
\newblock Entire Space Multi-Task Modeling via Post-Click Behavior
  Decomposition for Conversion Rate Prediction.
\newblock In \emph{Proceedings of the 43rd International {ACM} {SIGIR}
  conference on research and development in Information Retrieval}, 2377--2386.

\bibitem[{Yan et~al.(2014)Yan, Li, Xue, and Han}]{RelaImpr}
Yan, L.; Li, W.-J.; Xue, G.-R.; and Han, D. 2014.
\newblock Coupled group lasso for web-scale ctr prediction in display
  advertising.
\newblock In \emph{International Conference on Machine Learning}, 802--810.

\bibitem[{Yoshikawa and Imai(2018)}]{Nonpara_DFM}
Yoshikawa, Y.; and Imai, Y. 2018.
\newblock A Nonparametric Delayed Feedback Model for Conversion Rate Prediction
  abs/1802.00255.

\bibitem[{Zhou et~al.(2019)Zhou, Mou, Fan, Pi, Bian, Zhou, Zhu, and Gai}]{DIEN}
Zhou, G.; Mou, N.; Fan, Y.; Pi, Q.; Bian, W.; Zhou, C.; Zhu, X.; and Gai, K.
  2019.
\newblock Deep Interest Evolution Network for Click-Through Rate Prediction.
\newblock In \emph{The 33rd Association for the Advance of Artificial
  Intelligence Conference}, 5941--5948.

\bibitem[{Zhou et~al.(2018)Zhou, Zhu, Song, Fan, Zhu, Ma, Yan, Jin, Li, and
  Gai}]{DIN}
Zhou, G.; Zhu, X.; Song, C.; Fan, Y.; Zhu, H.; Ma, X.; Yan, Y.; Jin, J.; Li,
  H.; and Gai, K. 2018.
\newblock Deep Interest Network for Click-Through Rate Prediction.
\newblock In \emph{Proceedings of the 24th {ACM} {SIGKDD} International
  Conference on Knowledge Discovery {\&} Data Mining}, 1059--1068.

\bibitem[{Zhu et~al.(2017)Zhu, Jin, Tan, Pan, Zeng, Li, and Gai}]{GAUC2}
Zhu, H.; Jin, J.; Tan, C.; Pan, F.; Zeng, Y.; Li, H.; and Gai, K. 2017.
\newblock Optimized cost per click in taobao display advertising.
\newblock In \emph{Proceedings of the 23rd ACM SIGKDD International Conference
  on Knowledge Discovery and Data Mining}, 2191--2200.

\end{thebibliography}
	\end{document}